\documentclass[11pt]{article}

\usepackage[final]{acl}

\usepackage{times}
\usepackage{latexsym}

\usepackage[T1]{fontenc}

\usepackage[utf8]{inputenc}

\usepackage{microtype}

\usepackage{inconsolata}

\usepackage{graphicx}
\usepackage{amsmath}
\usepackage{algorithm}
\usepackage{algpseudocode}
\usepackage{multirow}
\usepackage[dvipsnames, svgnames, x11names]{xcolor}
\usepackage{colortbl}
\usepackage{float}
\usepackage{hyperref}
\definecolor{linkblue}{HTML}{000099}
\hypersetup{
    colorlinks=true,
    linkcolor=linkblue,
    urlcolor=linkblue,
    citecolor=linkblue,
    pdfborder={0 0 0}
}
\usepackage{amsmath, amssymb, amsthm}

\usepackage{enumitem}

\title{Learning Adaptive Parallel Execution for Efficient Code Localization}

\author{
  \textbf{Ke Xu\textsuperscript{1,2,*}},
  \textbf{Siyang Xiao\textsuperscript{1,*}},
  \textbf{Ming Liang\textsuperscript{1}},
  \textbf{Yichen Yu\textsuperscript{1}},
\\
  \textbf{Zhixiang Wang\textsuperscript{1,2}},
  \textbf{Jingxuan Xu\textsuperscript{1,3}},
  \textbf{Dajun Chen\textsuperscript{1}},
  \textbf{Wei Jiang\textsuperscript{1}},
  \textbf{Yong Li\textsuperscript{1,\dag}}
\\
\\
  \textsuperscript{1}Ant Group,
  \textsuperscript{2}Peking University,
  \textsuperscript{3}Beijing Jiaotong University
\\
  \small{
    \texttt{\{siyang.xsy, liangming.liang, yuyichen.yyc, chendajun.cdj, jonny.jw, liyong.liy\}@antgroup.com}
  }
\\
  \small{
    \texttt{\{xuke59, ekko\}@stu.pku.edu.cn,  23120315@bjtu.edu.cn}
  }
  \thanks{Equal contribution. Work done during the internship at Ant Group. \quad \textsuperscript{\dag}Corresponding author}
}

\begin{document}
\maketitle
\begin{abstract}
Code localization constitutes a key bottleneck in automated software development pipelines. While concurrent tool execution can enhance discovery speed, current agents demonstrate a 34.9\% redundant invocation rate, which negates parallelism benefits. We propose \textbf{FuseSearch}, reformulating parallel code localization as a \textbf{joint quality--efficiency optimization} task. Through defining \textbf{tool efficiency}---the ratio of unique information gain to invocation count---we utilize a two-phase SFT and RL training approach for learning adaptive parallel strategies. Different from fixed-breadth approaches, FuseSearch dynamically modulates search breadth according to task context, evolving from exploration phases to refinement stages. Evaluated on SWE-bench Verified, FuseSearch-4B achieves SOTA-level performance (84.7\% file-level and 56.4\% function-level $F_1$ scores) with 93.6\% speedup, utilizing 67.7\% fewer turns and 68.9\% fewer tokens. Results indicate that efficiency-aware training naturally improves quality through eliminating noisy redundant signals, enabling high-performance cost-effective localization agents.
\end{abstract}

\section{Introduction}
Code localization—identifying the relevant code entities needed to resolve a given issue—is a critical bottleneck in automated software development~\citep{jimenez2024swebench,xia2024agentlessdemystifyingllmbasedsoftware}. Recent studies show that state-of-the-art agents devote more than 50\% of their computational resources to this task, highlighting the need for more efficient strategies ~\citep{pan2025introducingswegrepandswegrepmini}. In response, recent work has proposed specialized localization agents that operate as dedicated search components, decoupling localization from downstream repair or generation steps~\citep{chen2025locagent,jiang2025issuelocalizationllmdriveniterative}. These agents typically rely on multi-turn interactions with \textbf{sequential tool execution} (e.g., code retrieval and analysis), iteratively refining queries, inspecting intermediate results, and narrowing down candidate files or functions to achieve high localization accuracy. However, this iterative paradigm introduces a fundamental trade-off: aggressive constraints on the number of allowed tool interactions (i.e., tight turn budgets) are increasingly necessary to meet real-world \textit{computational cost} requirements, as emphasized in recent benchmarks for production-grade agent systems~\citep{gao2025more}. Under such tight budgets, agents often fail to gather sufficient contextual evidence before exhausting their interaction quota—a phenomenon we refer to as information starvation. Consequently, further reductions in computational cost come at the cost of sharp accuracy degradation, limiting the deployability of current localization approaches in time-sensitive settings.

\begin{figure}[t]
  \centering
  \includegraphics[width=\columnwidth]{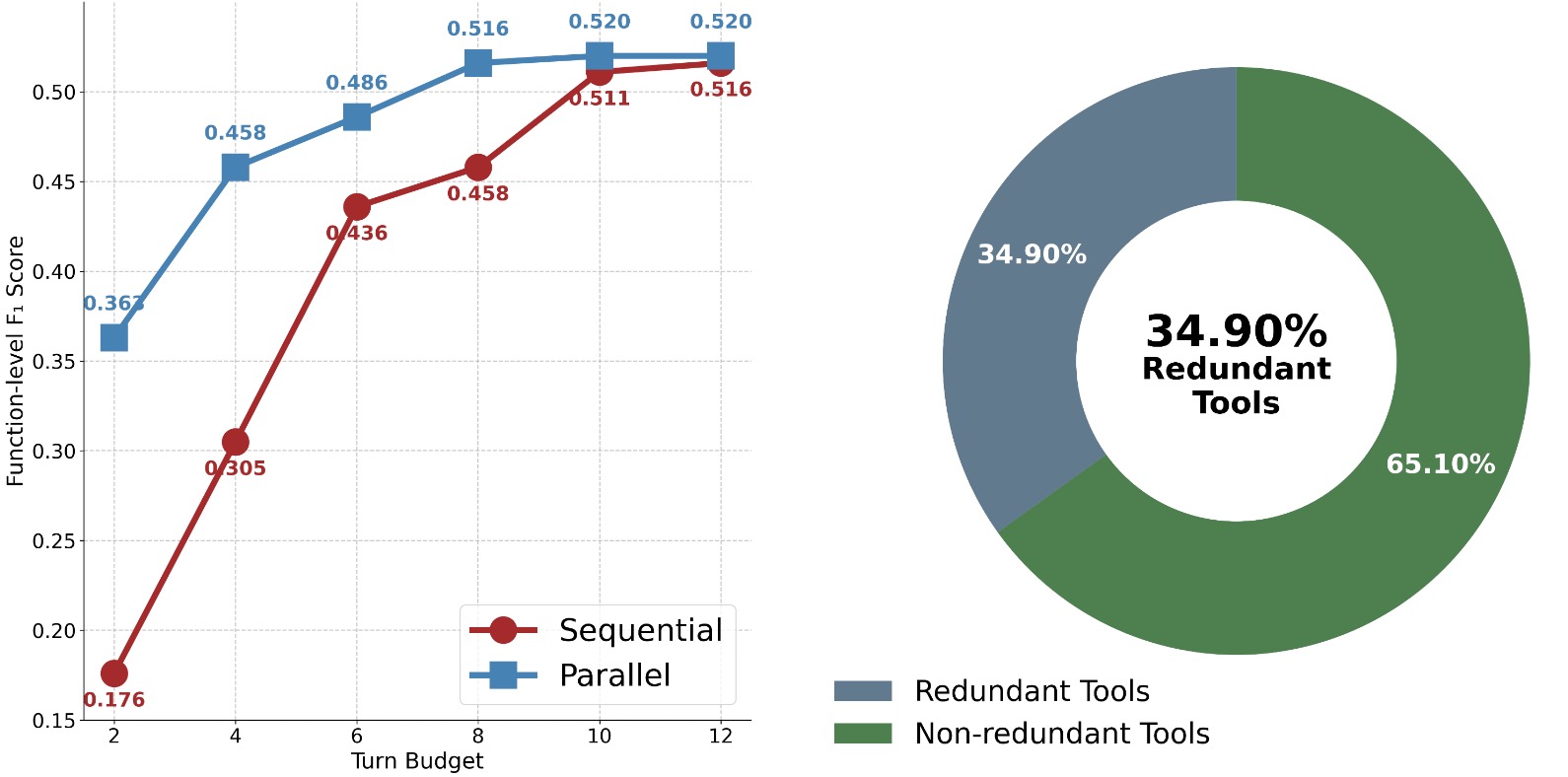}
    \caption{Parallel execution solves sequential search's information starvation under limited turns. However, 34.9\% of enforced parallel tools are redundant, exhibiting redundancy.}
    \label{fig:motivation}
    \vspace{-0.6cm}
\end{figure}

\textbf{Parallel tool execution} presents a promising avenue for addressing the cost–accuracy trade-off by enabling the simultaneous invocation of multiple tools within a single interaction turn, thereby increasing the information density per turn. As illustrated in Figure~\ref{fig:motivation}(a), under tight turn budgets, parallel execution significantly outperforms sequential search in terms of localization accuracy. 
Despite this potential, most existing agents only provide technical support for parallelism—allowing concurrent tool calls—without consistently harnessing its benefits in practice~\citep{pan2025introducingswegrepandswegrepmini}. 

Moreover, naive parallelization schemes that enforce a fixed number of tool calls per turn can be highly inefficient. As shown in Figure~\ref{fig:motivation}(b), such approaches incur more than 34.9\% redundant tool invocations. These unnecessary calls not only waste computational resources but also introduce noisy or irrelevant signals that can degrade localization performance.
This raises a key question: how can parallelization be made both \textbf{comprehensive} and \textbf{non-redundant}—\textit{maximizing information coverage} to avoid starvation under tight turn budgets while \textit{eliminating redundant exploration} of previously examined code? 



To address this challenge, we introduce \textbf{FuseSearch}, a code localization agent that achieves superior quality--efficiency trade-offs through \textit{learned adaptive parallel execution}. Instead of prescribing a fixed degree of parallelism, FuseSearch \textit{dynamically adjusts the breadth of parallel tool invocations} by explicitly optimizing \textbf{tool efficiency}—the ratio of tool calls that yield novel, relevant information to the total number of invocations. We instantiate tool efficiency as a reward that credits exploration of distinct code regions while penalizing redundancy and failed queries, enabling RL-based joint optimization of localization accuracy (measured by $F_1$) and search efficiency. FuseSearch adopts a minimalist design, relying only on three language-agnostic, read-only tools—\texttt{grep}, \texttt{glob}, and \texttt{read\_file}—and requires no auxiliary infrastructure such as code graphs or language-specific parsers.

Building on this formulation, we employ a two-stage training pipeline that combines SFT with RL to train compact models (4B and 30B parameters) to decide \textit{which} tools to invoke and \textit{how many} to run in parallel at each turn, balancing exploration breadth with precision. Notably, optimizing tool efficiency not only reduces search cost but also improves localization quality: by discouraging wasteful calls, the efficiency-driven reward guides the agent toward more targeted and accurate search strategies.

Experimental results on SWE-bench Verified~\citep{jimenez2024swebench} show that {FuseSearch (train)} substantially outperforms {RepoSearcher}~\citep{ma2025toolintegratedreinforcementlearningrepo} under the Qwen3-4B backbone ~\citep{qwen3technicalreport}. In terms of localization quality, FuseSearch delivers substantial gains, improving file-level $F_1$ from 38.1\% to 84.7\% and function-level $F_1$ from 21.7\% to 56.4\%, indicating markedly stronger precision in pinpointing both relevant files and target functions. Meanwhile, FuseSearch is significantly more efficient: it reduces overall interaction turns by {67.7\%}, cuts time by {93.6\%}, and lowers token consumption by {68.9\%}. These results suggest that the learned adaptive parallel execution not only boosts localization accuracy but also streamlines the search process, enabling faster and more targeted exploration with substantially less redundant tool usage. Our main contributions are:
\begin{itemize}
    \item We propose {tool efficiency} to quantify information novelty in code search and integrate it into an \textbf{efficiency-aware training framework} (via SFT and RL), enabling the joint optimization of search effectiveness and computational efficiency.
    \item We introduce {FuseSearch}, a minimalist localization agent that uses only three read-only tools (\texttt{grep}, \texttt{glob}, \texttt{read\_file}) yet matches or exceeds the performance of far more complex systems.
    \item We demonstrate that high-quality, low-latency localization significantly accelerates downstream agent workflows, cutting interaction turns by 23.1\% and end-to-end task time by 28.5\% without sacrificing success rates.
\end{itemize}

\section{Preliminary}
\label{sec:preliminary}

\subsection{Task Formulation}
Code localization aims to identify the specific code entities—such as files, functions, or code snippets—that require modification to resolve a given issue. We formulate this problem as a repository-level \emph{information-seeking task}. Unlike static retrieval approaches, this process involves an agent that actively interacts with the repository to progressively accumulate relevant context. Through iterative tool calls, the agent retrieves and analyzes various code entities, narrowing down the search space to produce the final localization result.

Formally, an agent operates over discrete turns $t = 1, \ldots, T$. A search trajectory is defined as:
\begin{equation}
\tau = (q, a_1, o_1, \ldots, a_T, o_T, \mathcal{A})
\end{equation}
where $q$ is the issue description, $a_t$ is the set of tool calls at turn $t$, $o_t$ is the aggregated observation containing the retrieved code content, and $\mathcal{A}$ is the final localization result identifying the target entities for modification.

\subsection{Parallel Tool Execution}
Traditional sequential agents invoke one tool per turn, leading to prolonged search duration when comprehensive exploration is needed. Parallel tool execution enables simultaneous invocation of multiple tools within a single turn, increasing information density per interaction. In this paradigm, agents generate multiple tool calls in one response, each formatted as a JSON object. All tools within a turn execute concurrently—their read-only nature eliminating synchronization concerns—and their results are aggregated before the next agent response. This design reduces the total number of interaction turns required and shortens overall search time.

\section{FuseSearch}

\begin{figure*}[t]
  \centering
  \includegraphics[width=\textwidth]{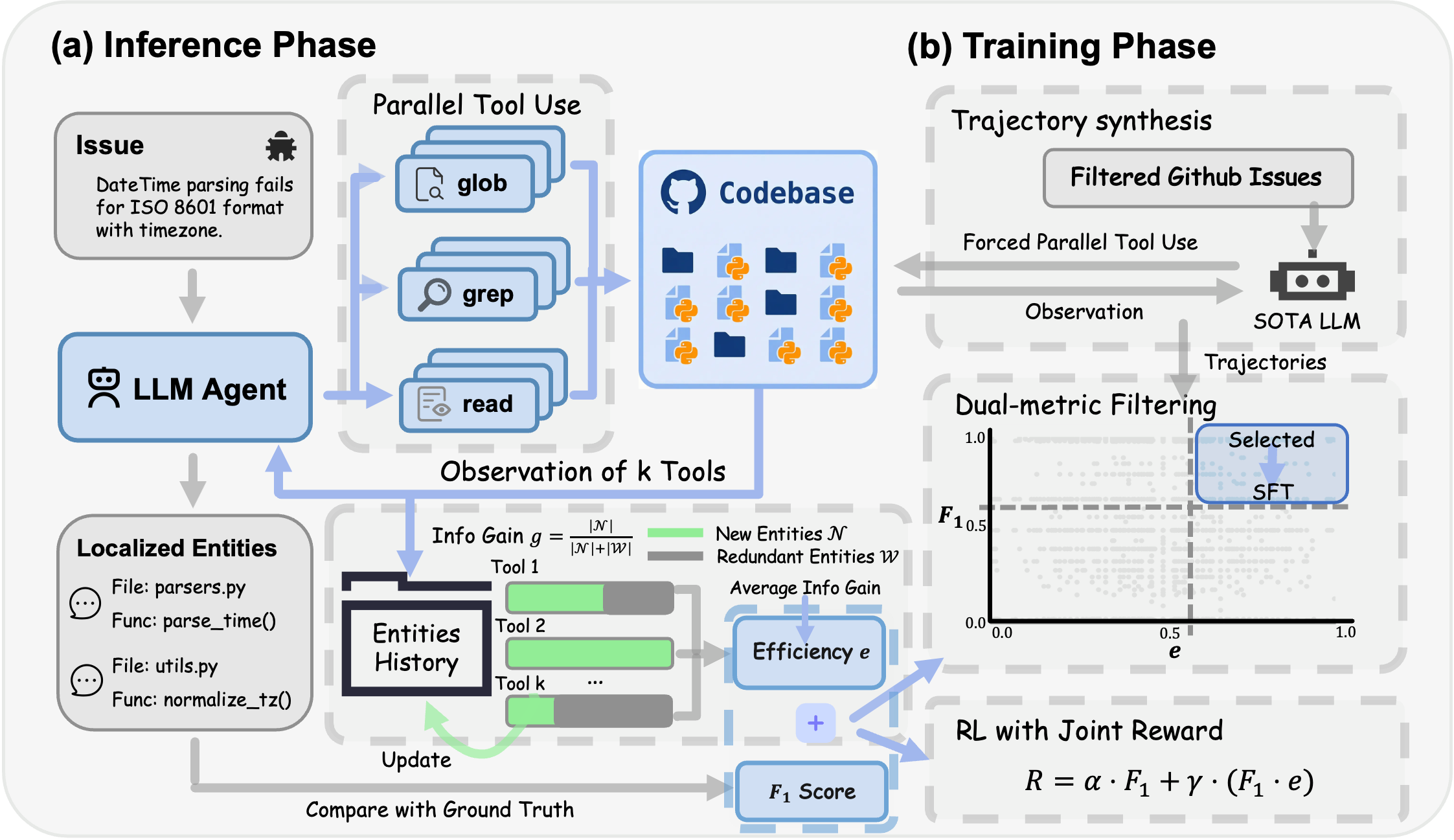}
\caption{FuseSearch framework overview. (a) \textbf{Inference}: Agent executes three minimalist tools in parallel, with each tool's information gain tracked to compute trajectory efficiency $e$. (b) \textbf{Training}: Dual-metric filtering selects high-quality trajectories for SFT, followed by RL optimization with joint $F_1$-efficiency reward.}
  \label{fig:framework}
  \vspace{-0.5cm}
\end{figure*}

We present FuseSearch, a minimalist framework for efficient code localization through learned parallel execution. This section is organized as follows: Section~\ref{sec:toolset} introduces our minimalist tool set. Section~\ref{sec:efficiency_metrics} defines efficiency metrics and dual-objective optimization framework—the key innovation enabling joint quality-efficiency optimization. Section~\ref{sec:training} details the training approach implementing these objectives. Figure~\ref{fig:framework} illustrates the overall architecture.

\subsection{Minimalist Tool Set}
\label{sec:toolset}
We employ a minimalist architecture comprising three read-only tools that enable effective code localization without infrastructure dependencies:

\begin{itemize}[leftmargin=*, topsep=2pt, itemsep=1pt]
    \item \texttt{grep}: Regex-based pattern search in file contents
    \item \texttt{glob}: File path pattern matching
    \item \texttt{read\_file}: Reading file contents with optional line range specification
\end{itemize}

This minimalist design offers several practical advantages. First, the language-agnostic nature requires no parsers or runtime environments, enabling immediate deployment across diverse codebases. Second, the simplicity reduces learning complexity for models, allowing training resources to focus on strategic tool orchestration rather than intricate tool semantics. Third, it eliminates dependency on pre-computed structures like ASTs or dependency graphs, avoiding the overhead of building and maintaining auxiliary infrastructure.

\subsection{Dual-Objective Metrics for Code Localization}
\label{sec:efficiency_metrics}

While parallel tool execution increases information throughput, naive parallelization often results in substantial redundancy. As shown in Figure~\ref{fig:motivation}, over 34.9\% of tools in enforced parallel execution provide no incremental value. To optimize parallel search strategies, we must quantify search effectiveness along two complementary dimensions: output quality and process efficiency. We define metrics for both dimensions and establish their joint optimization framework.

\paragraph{Localization Quality}
Following standard practice~\citep{xia2024agentlessdemystifyingllmbasedsoftware,chen2025locagent}, we measure localization quality using precision $\mathcal{P}$, recall $\mathcal{R}$, and their harmonic mean $F_1$ at file-level and function-level granularities. Let $\hat{\mathcal{A}}$ denote the predicted entity set and $\mathcal{A}$ the ground truth:
\begin{equation}
\mathcal{P} = \frac{|\hat{\mathcal{A}} \cap \mathcal{A}|}{|\hat{\mathcal{A}}|}, \quad
\mathcal{R} = \frac{|\hat{\mathcal{A}} \cap \mathcal{A}|}{|\mathcal{A}|}, \quad
F_1 = \frac{2\mathcal{P}\mathcal{R}}{\mathcal{P} + \mathcal{R}}
\end{equation}

The $F_1$ score balances precision and recall, rewarding models that identify relevant code comprehensively (high recall) without excessive over-prediction (high precision).

\paragraph{Tool Efficiency}
To quantify search efficiency, we measure the information gain of each tool call relative to search progress. During execution, we maintain a history of discovered code entities, including files accessed and content regions examined. For each tool call, we compute its \emph{information gain} $g_i$ by comparing returned entities against this cumulative history:

\begin{equation}
    g_i = 
    \begin{cases}
        \frac{|\mathcal{E}_i \setminus \mathcal{H}|}{|\mathcal{E}_i|} & \text{if } |\mathcal{E}_i| > 0 \\
        0 & \text{otherwise}
    \end{cases}
\end{equation}
where $\mathcal{E}_i$ denotes the set of entities returned by tool $i$, and $\mathcal{H}$ represents the union of all entities discovered in preceding turns. The term $\mathcal{E}_i \setminus \mathcal{H}$ thus quantifies the incremental knowledge gain provided by the tool.

Tool calls exhibit diverse effectiveness patterns. Some discover entirely new content ($g_i = 1.0$), others retrieve only previously-seen entities ($g_i = 0$), and many return mixed results with partial overlap ($0 < g_i < 1$). We define tool efficiency $e$ as the mean information gain across all tool invocations:

\begin{equation}
e = \frac{1}{k} \sum_{i=1}^{k} g_i
\end{equation}
where $k$ is the total number of tool calls. This metric credits tools proportionally to their novel contributions: higher $e$ indicates exploration discovering distinct code regions, while lower $e$ reveals wasteful redundancy. 

\paragraph{Joint Optimization Objectives}
\label{subsec:objectives}

We formulate the agent learning objective as jointly maximizing localization quality $F_1$ and tool efficiency $e$.

This joint objective induces mutually reinforcing optimization dynamics. High $F_1$ requires collecting sufficient yet focused information: insufficient exploration yields incomplete coverage, while excessive unfocused exploration creates context overload that degrades answer precision. High $e$ requires each tool to explore distinct regions of the search space, avoiding redundant queries.

These objectives are complementary rather than conflicting. An agent cannot achieve high $e$ through low-information tools—empty results or duplicated observations yield zero information gain. However, high $e$ alone is insufficient: brute-force strategies like sequentially reading distinct but irrelevant files achieve high efficiency (each file is novel) yet produce low $F_1$, as context accumulation with off-target exploration degrades answer precision. The joint objective thus enforces \emph{focused, non-redundant exploration}: agents must efficiently locate relevant code without wasteful detours.

\subsection{Joint Quality-Efficiency Training}
\label{sec:training}
To optimize the dual objectives defined in Section~\ref{subsec:objectives}—achieving high localization quality $F_1$ and search efficiency $e$—we employ a two-stage training approach. SFT provides initial capabilities for parallel tool execution and establishes a strong baseline for both metrics. RL then refines the policy to jointly maximize $F_1$ and $e$ through strategic exploration.

\paragraph{Training Data Construction} 
We construct a repository-level code localization dataset from 233 high-quality GitHub repositories, ensuring no overlap with our evaluation benchmarks. To ensure data quality, we exclude samples where (1) patches introduce entirely new files or functions, (2) issue descriptions are incomplete or overly brief, or (3) no code changes occur. From $\sim$21K filtered samples, we extract ground truth localization targets as the modified files, functions/methods, and line ranges from each patch.

\paragraph{Quality-Efficiency Guided Fine-Tuning}

Base language models exhibit weak parallel tool usage capabilities, occasionally generating at most a few tool calls per turn. To bootstrap reliable parallel execution while ensuring high initial quality and efficiency, we perform SFT on trajectories filtered by both $F_1$ and $e$ metrics.

We use a capable teacher model (Kimi-K2-Instruct) to synthesize training trajectories. Since even advanced models exhibit inconsistent parallel behavior, we employ system-level guidance to increase parallel execution frequency: for 6K randomly sampled training queries, we generate multiple trajectories per query, each explicitly guided to use 2-8 tools per turn, yielding approximately 24K candidate trajectories. We then apply dual-metric filtering, retaining only trajectories satisfying:
\begin{equation}
F_1 \geq \rho_F \quad \text{and} \quad e \geq \rho_e
\end{equation}
This filtering ensures demonstration data exhibits both accurate localization and high tool efficiency. The resulting $\sim$6K high-quality trajectories are used for fine-tuning Qwen3 base models (4B and 30B-A3B).

The resulting SFT models serve dual purposes: (1) they reliably generate parallel tool calls (2-8 tools per turn), addressing the base model's limited parallel execution capability, and (2) they provide high-quality initialization with reasonable $F_1$ and $e$ values, enabling effective trajectory sampling during subsequent RL training.

\paragraph{RL with Joint Reward}

Building on the SFT initialization, we apply group relative policy optimization (GRPO)~\citep{shao2024deepseekmathpushinglimitsmathematical} to further optimize both localization quality and search efficiency. GRPO samples multiple outputs per query, computes advantages based on reward signals, and updates the policy to favor high-reward behaviors while maintaining proximity to a reference policy through KL regularization.

To jointly optimize localization quality ($F_1$) and search efficiency ($e$), we consider a general reward function encompassing both linear and interactive contributions:
\begin{equation}
    R(\tau) = \alpha \cdot F_1(\tau) + \beta \cdot e(\tau) + \gamma \cdot (F_1(\tau) \cdot e(\tau))
\end{equation}
where $\alpha, \beta, \gamma \geq 0$ are weighting coefficients.

For code localization, we impose a strict boundary condition: a trajectory that fails to identify relevant code ($F_1=0$) provides zero utility, regardless of how "efficiently" it executed. This constraint necessitates setting $\beta = 0$, yielding:
\begin{equation}
    R(\tau) = \underbrace{\alpha \cdot F_1(\tau)}_{\text{Base Guarantee}} + \underbrace{\gamma \cdot (F_1(\tau) \cdot e(\tau))}_{\text{Efficiency Bonus}}
\end{equation}
The linear term guarantees a baseline reward for correct localization, preventing the vanishing reward problem when efficiency is low. The interaction term acts as a soft gate, amplifying the reward only when high quality is achieved with high efficiency.

In practice, $F_1$ is computed as a weighted combination of file-level and function-level localization accuracy:
\begin{equation}
F_1 = \lambda_{\text{file}} \cdot F_1^{\text{file}} + \lambda_{\text{func}} \cdot F_1^{\text{func}}
\end{equation}
where $F_1^{\text{file}}$ and $F_1^{\text{func}}$ measure precision and recall at their respective granularities. The efficiency metric $e$ is computed as defined in Section~\ref{sec:efficiency_metrics}. By explicitly coupling search efficiency with localization quality, this objective aligns the RL signal with our dual goals, encouraging the model to maximize information gain per action without compromising the validity of the final result.

\section{Experiments}
\label{sec:exp}

\paragraph{Datasets} We evaluate on SWE-bench Verified~\citep{jimenez2024swebench}, a curated benchmark for repository-level issue resolution. Following~\citet{suresh2025cornstackhighqualitycontrastivedata}, we exclude examples where patches introduce entirely new files or functions, retaining 386 of 500 examples.

\paragraph{Metrics} We evaluate localization quality using precision, recall, and $F_{1}$ scores at both file-level and function-level granularities. We measure search cost through wall-clock time ($T(s)$), interaction turns (\#Turn), and total tokens consumed (Tok.(k)) per instance, capturing both latency and computational overhead (averaged over three runs).

\paragraph{Baselines} We compare against three categories: (1) \textbf{Workflow-based}: Agentless~\citep{xia2024agentlessdemystifyingllmbasedsoftware}; (2) \textbf{Agent-based}: LocAgent~\citep{chen2025locagent}, CoSIL~\citep{jiang2025issuelocalizationllmdriveniterative}, and RepoSearcher~\citep{ma2025toolintegratedreinforcementlearningrepo}. Implementation details are provided in Appendix~\ref{sec:baseline_implementation}.

\begin{table*}[t]
  \centering
  \small
  \renewcommand{\arraystretch}{1.2}
  \begin{tabular}{cc|ccc|ccc|c@{\hspace{0.3cm}}c@{\hspace{0.2cm}}c|c}
    \hline
    \multirow{2}{*}{\textbf{Model}} & \multirow{2}{*}{\textbf{Method}} & \multicolumn{3}{c|}{\textbf{File (\%)}} & \multicolumn{3}{c|}{\textbf{Func (\%)}} & \multirow{2}{*}{\textbf{\#Turn}} & \multirow{2}{*}{\textbf{T(s)}} & \multirow{2}{*}{\textbf{Tok.(k)}} & \multirow{2}{*}{\textbf{Mode}} \\
    \cline{3-8}
    & & $P$ & $R$ & $F_{1}$ & $P$ & $R$ & $F_{1}$ & & & & \\
    \hline\hline
    \rowcolor{gray!20} \multicolumn{12}{c}{\textbf{Proprietary Models}} \\
    \hline
    \multirow{6}{*}{Haiku 4.5} & Agentless & 38.82 & 91.71 & 54.55 & 21.48 & 61.37 & 31.83 & 2.00 & 7.32 & 10.6 & Seq \\
    \cline{2-12}
    & CoSIL & 19.62 & 96.63 & 32.62 & 18.37 & 69.41 & 29.05 & 7.22 & 38.7 & \textbf{53.0} & Seq \\
    & LocAgent & 61.57 & 87.56 & 72.30 & 41.29 & 65.49 & 50.64 & 17.3 & 318 & 567 & Seq \\
    & RepoSearcher & 19.55 & 97.41 & 32.57 & 20.33 & 67.65 & 31.26 & 19.7 & 114 & 193 & Seq \\
    & FuseSearch* & 86.38 & 62.44 & 72.48 & 66.30 & 47.45 & 55.31 & 23.9 & 90.3 & 270 & Seq \\
    \rowcolor{MediumPurple!10}
    \cellcolor{white} & FuseSearch & 73.54 & 94.50 & \textbf{82.71} & 48.58 & 70.61 & \textbf{57.56} & \textbf{6.24} & \textbf{36.2} & 110 & Par \\
    \hline
    \rowcolor{gray!20} \multicolumn{12}{c}{\textbf{Open-Source Models}} \\
    \hline
    \multirow{6}{*}{Kimi-K2} & Agentless & 34.53 & 92.23 & 50.25 & 28.22 & 56.27 & 37.59 & 2.00 & 11.8 & 8.33 & Seq \\
    \cline{2-12}
    & CoSIL & 21.33 & 95.60 & 34.88 & 23.04 & 70.20 & 34.69 & 6.8 & 58.8 & 94.0 & Seq \\
    & LocAgent & 55.39 & 95.85 & 70.21 & 33.10 & 73.33 & 45.61 & 14.9 & 261 & 447 & Seq \\
    & RepoSearcher & 20.61 & 96.11 & 33.94 & 25.20 & 72.35 & 37.39 & 15.6 & 94.5 & 108 & Seq \\
    & FuseSearch* & 77.33 & 79.53 & 78.42 & 51.87 & 49.02 & 50.40 & 15.0 & 71.3 & 216 & Seq \\
    \rowcolor{MediumPurple!10}
    \cellcolor{white} & FuseSearch & 75.11 & 89.31 & \textbf{81.60} & 51.00 & 54.90 & \textbf{52.88} & \textbf{7.92} & \textbf{43.6} & \textbf{62.1} & Par \\
    \hline
    \multirow{6}{*}{Qwen3-4B} & Agentless & 28.11 & 76.68 & 41.14 & 11.62 & 34.12 & 17.33 & 2.00 & 4.24 & 8.33 & Seq \\
    \cline{2-12}
    & CoSIL & 21.21 & 94.82 & 34.66 & 18.17 & 65.49 & 28.45 & 10.8 & 50.8 & 63.9 & Seq \\
    & LocAgent & 39.10 & 56.22 & 46.12 & 27.37 & 31.18 & 26.97 & 6.09 & 109 & 135 & Seq \\
    & RepoSearcher & 31.99 & 47.15 & 38.12 & 16.92 & 30.39 & 21.74 & 14.8 & 85.3 & 99.2 & Seq \\
    & FuseSearch (base) & 64.75 & 64.25 & 64.50 & 43.95 & 34.90 & 38.91 & \textbf{4.24} & 6.12 & 47.9 & Par \\
    \rowcolor{MediumPurple!25}
    \cellcolor{white} & FuseSearch (train) & 83.59 & 85.75 & \textbf{84.65} & 59.91 & 53.33 & \textbf{56.43} & 4.78 & \textbf{5.43} & \textbf{30.9} & Par \\
    \hline
    \multirow{6}{*}{Qwen3-30B} & Agentless & 24.22 & 87.56 & 38.19 & 15.34 & 55.10 & 24.00 & 2.00 & 8.04 & 32.0 & Seq \\
    \cline{2-12}
    & CoSIL & 21.22 & 96.11 & 34.77 & 18.40 & 66.86 & 28.86 & 11.1 & 49.2 & 66.8 & Seq \\
    & LocAgent & 45.16 & 62.95 & 52.59 & 29.96 & 32.75 & 31.32 & 11.4 & 112 & 136 & Seq \\
    & RepoSearcher & 40.25 & 50.78 & 44.90 & 20.07 & 34.71 & 25.43 & 16.7 & 92.4 & 113 & Seq \\
    & FuseSearch (base) & 70.41 & 79.53 & 74.70 & 53.27 & 46.27 & 45.65 & 7.50 & 14.9 & 80.1 & Par \\
    \rowcolor{MediumPurple!25}
    \cellcolor{white} & FuseSearch (train) & 83.12 & 82.90 & \textbf{83.01} & 66.58 & 52.35 & \textbf{58.62} & \textbf{5.77} & \textbf{10.6} & \textbf{43.2} & Par \\
    \hline
  \end{tabular}
  \caption{Localization performance and efficiency comparison on SWE-bench Verified. For agent-based and workflow-based methods, we evaluate using Kimi-K2-Instruct (abbr. as Kimi-K2) and Claude Haiku 4.5 (abbr. as Haiku 4.5). For FuseSearch, we compare base Qwen3 models (Qwen3-4B-Instruct and Qwen3-30B-A3B-Instruct) with their trained counterparts. FuseSearch* denotes using the FuseSearch framework with sequential prompts to contrast the two execution modes.}
  \label{tab:main_results}
  \vspace{-0.4cm}
\end{table*}

\subsection{Overall Performance}
\label{sec:main_results}

Table~\ref{tab:main_results} presents our main results on SWE-bench Verified. We highlight three key findings:

\paragraph{Parallel vs. Sequential Execution} 
Comparing sequential and parallel execution modes with identical toolsets, parallel invocation achieves comparable or superior localization quality while significantly reducing search time (e.g., ~60\% on Haiku 4.5) and requiring substantially fewer interaction turns. This demonstrates that parallelization provides efficiency gains beyond mere latency reduction—simultaneous information gathering enables better-informed decisions at each search step.
\paragraph{Minimalist Toolset Effectiveness} 
Even in sequential mode, our minimalist toolset achieves competitive performance compared to specialized agent-based methods with graph navigation or AST parsing. This indicates that language-agnostic primitives suffice for effective code localization, while being simpler to deploy and maintain.
\paragraph{Training Effects} 
Targeted training with joint $F_1$ and efficiency optimization substantially improves both precision and recall. Our trained Qwen3 models (4B and 30B) achieve 83-84\% file F$_1$ and 56-58\% function $F_1$, matching Claude Haiku 4.5's performance while being significantly faster and more token-efficient. Additional evaluation on LocBench~\citep{chen2025locagent} further confirms the superior performance of our trained models (Appendix~\ref{sec:additional_experiments}).

\begin{table}[t]
  \centering
  \small
  \renewcommand{\arraystretch}{1.2}
  \begin{tabular}{@{}l@{\hspace{2pt}}c@{\hspace{2pt}}c@{\hspace{2pt}}c@{\hspace{4pt}}c@{\hspace{3pt}}c@{\hspace{4pt}}c@{\hspace{3pt}}c@{}}
    \hline
    \textbf{Stage} & \textbf{File F$_1$} & \textbf{Func F$_1$} & \textbf{e} & \textbf{\#Turn} & \textbf{\#Tool} & \textbf{T(s)} & \textbf{Tok.(k)} \\
    \hline\hline
    \rowcolor{gray!20} \multicolumn{8}{c}{\textbf{Qwen3-4B}} \\
    \hline
    Base & 64.50 & 38.91 & 59.50 & 4.24 & 1.63 & 6.12 & 47.9 \\
    RL & 70.11 & 40.18 & 54.01 & \textbf{3.16} & 3.44 & 7.10 & 31.7 \\
    SFT & 78.86 & 47.94 & 68.46 & 4.96 & 3.59 & 9.17 & 54.8 \\
    SFT+RL & \textbf{84.65} & \textbf{56.43} & \textbf{69.00} & 4.78 & 2.15 & \textbf{5.43} & \textbf{30.9} \\
    \hline
    \rowcolor{gray!20} \multicolumn{8}{c}{\textbf{Qwen3-30B-A3B}} \\
    \hline
    Base & 74.70 & 45.65 & 54.92 & 7.50 & 1.65 & 14.9 & 80.1 \\
    RL & 79.17 & 47.67 & 49.21 & \textbf{4.23} & 1.24 & 6.63 & 51.0 \\
    SFT & 81.13 & 51.17 & 59.80 & 5.49 & 3.40 & 11.7 & 65.2 \\
    SFT+RL & \textbf{83.01} & \textbf{58.62} & \textbf{64.53} & 5.77 & 3.44 & \textbf{10.6} & \textbf{43.2} \\
    \hline
  \end{tabular}
  \caption{Progressive training effects. SFT establishes parallel tool usage from high-quality trajectories, while RL refines search strategies through F$_1$ optimization.}
  \label{tab:training_stages}
\vspace{-0.6cm}
\end{table}

\subsection{Training Analysis}
\label{sec:training_analysis}

Table~\ref{tab:training_stages} reveals key insights about our training framework.

\paragraph{Complementary Training Stages} 
The two-stage training exhibits clear complementarity. SFT establishes parallel tool usage capability and substantially improves $F_1$, but introduces redundancy that degrades efficiency and increases search cost. RL then resolves this trade-off through joint optimization: it further improves $F_1$ while simultaneously recovering efficiency and reducing time cost. This validates our design—neither stage alone achieves optimal quality-efficiency balance, but their combination enables effective parallelization without sacrificing precision.

\begin{figure}[t]
  \centering
  \includegraphics[width=\columnwidth]{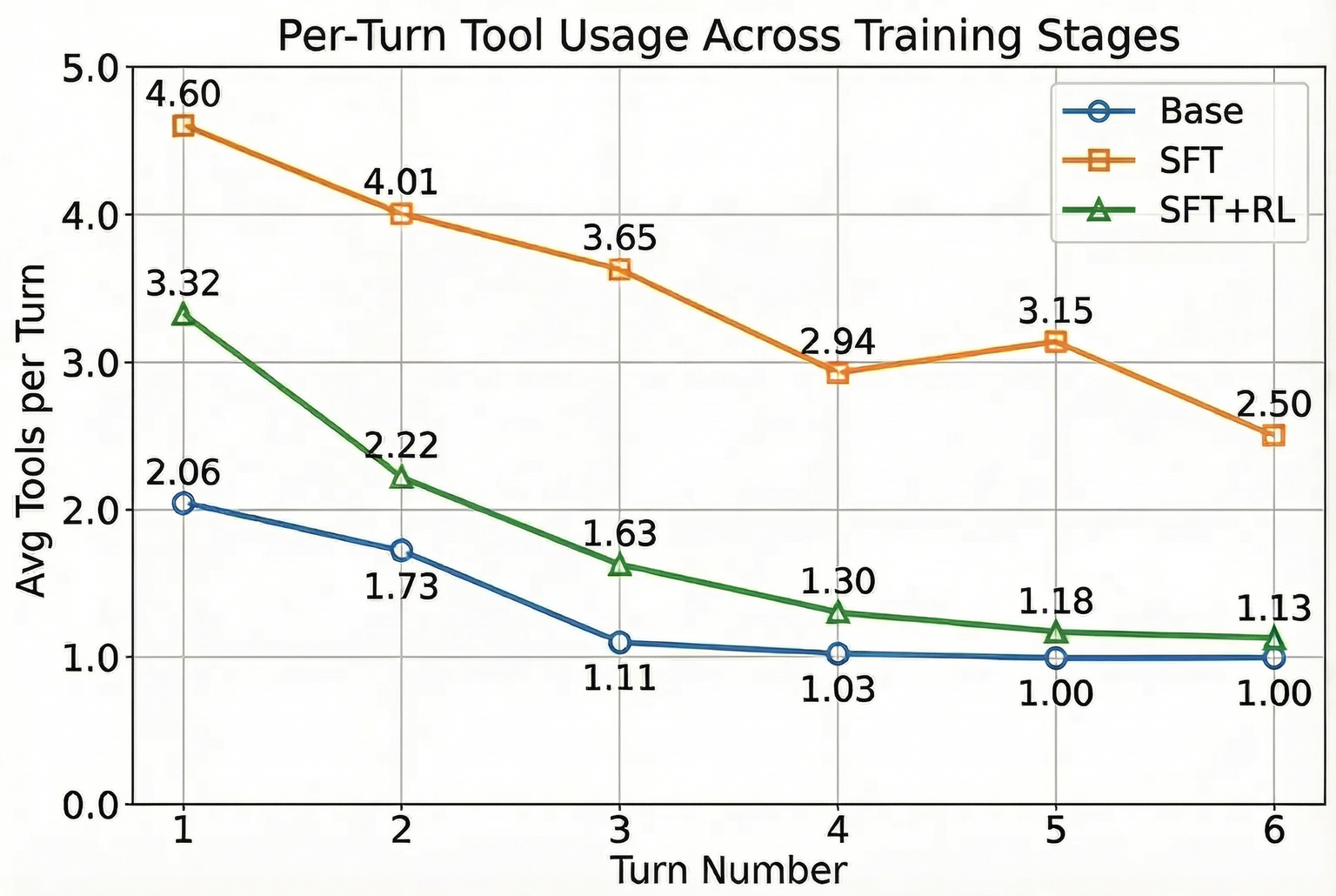}
    \caption{Evolution of average tools per turn across training stages. RL learns adaptive parallelism: high initial exploration transitioning to focused refinement.}
    \label{fig:per_turn_tools}
    \vspace{-0.6cm}
\end{figure}

\paragraph{Evolving Parallel Strategies} Figure~\ref{fig:per_turn_tools} reveals how parallel search strategies evolve across training. Base models use minimal parallelism; SFT shifts to uniformly aggressive parallelism, explaining both recall gains and efficiency drops. RL produces qualitatively different behavior: adaptive parallelism that begins with broad exploration and rapidly transitions to focused refinement. This breadth-first-to-depth-first pattern emerges from joint optimization, demonstrating that models learn not just to parallelize, but when and how much.

\subsection{Ablation Studies}
\label{sec:ablation}

We validate key design choices through systematic ablations on Qwen3-4B.
\paragraph{Parallel vs. Sequential Execution}
To isolate execution mode impact, we train sequential variants (1 tool/turn) with identical data and configurations. Table~\ref{tab:parallel_ablation} shows parallel execution consistently outperforms sequential. Sequential requires nearly twice as many turns yet achieves lower F$_1$ and consumes more tokens. This confirms parallel execution offers fundamental advantages beyond latency—simultaneous information gathering enables better decisions at each step.

\paragraph{SFT Data Filtering}

We evaluate four filtering strategies, each yielding 6K trajectories for SFT: (A) no filtering, (B) F$_1$-only filtering, (C) efficiency-only filtering, (D) joint filtering (ours). Table~\ref{tab:filtering_ablation} shows joint filtering produces SFT models with superior F$_1$ and efficiency simultaneously. Single-metric filtering improves its target metric but degrades the other, while unfiltered data yields suboptimal initialization on both. This validates that dual-metric filtering provides high-quality starting points for subsequent RL optimization.

\begin{table}[t]
  \centering
  \small
  \setlength{\tabcolsep}{2.5pt} 
  \renewcommand{\arraystretch}{1.3}
  \begin{tabular}{lcccccc}
    \hline
    \textbf{Mode} & \textbf{Stage} & \textbf{File F$_1$} & \textbf{Func F$_1$} & \textbf{\#Turn} & \textbf{T(s)} & \textbf{Tok.(k)}\\
    \hline
    \multirow{2}{*}{Seq} & SFT & 74.02 & 47.15 & 9.82 & 10.42 & 95.9 \\
    & SFT+RL & 78.82 & 50.21 & 7.52 & 8.03 & 59.4 \\
    \hline
    \multirow{2}{*}{Par} & SFT & 78.86 & 47.94 & 4.96 & 9.17 & 54.8 \\
    & SFT+RL & 84.65 & 56.45 & 5.60 & 5.43 & 30.9 \\
    \hline
  \end{tabular}
  \caption{Parallel vs sequential execution with identical training configurations. Parallel consistently outperforms sequential across both training stages.}
  \label{tab:parallel_ablation}
\vspace{-0.3cm}
\end{table}

\begin{table}[t]
  \centering
  \small
  \setlength{\tabcolsep}{3.5pt} 
  \renewcommand{\arraystretch}{1.3}
  \begin{tabular}{lccccc}
    \hline
    \textbf{Filtering} & \textbf{File F$_1$} & \textbf{Func F$_1$} & \textbf{e} & \textbf{T(s)} & {\textbf{Tok.(k)}} \\
    \hline
    No filtering & 75.44 & 43.52 & 55.77 & 9.84 & 60.7  \\
    Filter $F_1$ & 78.55 & 45.43 & 56.72 & 10.53 & 73.2 \\
    Filter $e$ & 76.74 & 42.63 & 60.14 & 12.94 & 61.8 \\
    Joint filtering & 78.86 & 47.94 & 62.03 & 9.17 & 54.8 \\
    \hline
  \end{tabular}
  \caption{SFT performance under different filtering strategies. Joint filtering achieves optimal F$_1$ and efficiency simultaneously.}
  \label{tab:filtering_ablation}
\vspace{-0.3cm}
\end{table}

\begin{table}[t]
  \centering
  \small
  \setlength{\tabcolsep}{3pt} 
  \renewcommand{\arraystretch}{1.3}
  \begin{tabular}{lccccc}
    \hline
    \textbf{Reward Type} & \textbf{File F$_1$} & \textbf{Func F$_1$} & \textbf{e} & \textbf{T(s)} & \textbf{Tok.(k)} \\
    \hline
    SFT & 78.86 & 47.94 & 62.03 & 9.17 & 54.8 \\
    \hline
    $F_1$ only & 81.84 & 54.90 & 59.66 & 7.28 & 39.4 \\
    $F_1$ + $e$ & 79.22 & 51.98 & 66.62 & 9.40 & 45.7 \\
    $F_1+F_1 \cdot e$ (ours) & 84.65 & 56.45 & 69.00 & 5.43 & 30.9 \\
    \hline
  \end{tabular}
  \caption{Reward design ablation. The composite reward balances quality and efficiency optimally.}
  \label{tab:reward_ablation}
\vspace{-0.5cm}
\end{table}

\paragraph{Reward Design}
We compare three RL reward formulations: (A) $F_1$ only, (B) additive $F_1 + e$, (C) multiplicative $F_1+F_1 \cdot e$ (ours). Table~\ref{tab:reward_ablation} reveals distinct trade-offs. $F_1$-only optimization improves over SFT with reduced token usage but moderate time reduction. Additive $F_1 + e$ achieves highest efficiency, but reduced search quality necessitates more tool invocations, increasing both time and token costs. Our multiplicative reward delivers the best $F_1$, highest efficiency, and lowest time and token costs simultaneously. This validates the multiplicative reward structure for practical deployment scenarios where both quality and efficiency matter.

\subsection{Downstream Task Applications}

To evaluate FuseSearch's practical impact, we test its effectiveness in assisting Kimi-K2-Instruct on SWE-bench Verified issue resolution. We compare three configurations: (A) \textbf{No Localization}, where the main agent performs full-stack exploration; (B) \textbf{Pre-Search}, where FuseSearch-4B conducts initial localization before the main agent begins; and (C) \textbf{Sub-Agent}, where the main agent dynamically invokes FuseSearch-4B during task execution.

Table~\ref{tab:downstream} shows that both localization modes maintain comparable pass rates while substantially reducing the main agent's token consumption and total inference time. Pre-search mode offers the most efficient configuration, demonstrating that fast, accurate localization models can significantly accelerate downstream task completion without sacrificing solution quality.

\begin{table}[t]
  \centering
  \small
  \setlength{\tabcolsep}{4pt}
  \renewcommand{\arraystretch}{1.3}
  \begin{tabular}{lcccc}
    \hline
    \textbf{Method} & \textbf{Pass Rate (\%)} & \textbf{\#Turn} & \textbf{T(s)} & \textbf{Tok.(k)} \\
    \hline
    No Localization & 68.4 & 41.1 & 312 & 1053 \\
    Pre-Search & 68.1 & 31.6 & 223 & 562 \\
    Sub-Agent & 68.7 & 31.9 & 290 & 713 \\
    \hline
  \end{tabular}
  \caption{Impact of FuseSearch-4B on Kimi-K2's issue resolution performance. Localization reduces token consumption and time while maintaining quality.}
  \label{tab:downstream}
\vspace{-0.6cm}
\end{table}

\section{Related Work}
\subsection{Code Localization Methods}

Recent LLM-based approaches divide into two paradigms. \textbf{Workflow methods} like Agentless~\citep{xia2024agentlessdemystifyingllmbasedsoftware} employ fixed hierarchical strategies that progressively narrow search scope from files to functions, but lack adaptability to varying task complexity. \textbf{Agent-based methods} enable flexible multi-step exploration: graph-guided approaches like LocAgent~\citep{chen2025locagent} and CoSIL~\citep{jiang2025issuelocalizationllmdriveniterative} represent code as static or dynamic graphs for navigation, while general agents like OpenHands~\citep{wang2025openhandsopenplatformai} and SWE-agent~\citep{yang2024sweagentagentcomputerinterfacesenable} use bash-like interfaces for repository traversal.

However, existing agents execute tool calls sequentially, leading to prolonged search duration when comprehensive exploration is needed. Graph-based methods further require language-specific preprocessing infrastructure that limits generalization across programming languages. We address these limitations through a minimalist parallel execution framework that accelerates search via concurrent tool invocation while eliminating infrastructure dependencies.
\subsection{Parallel Tool Use and Agent Training}

Parallel tool execution has emerged as a strategy to accelerate multi-step search processes. Commercial systems like SWE-grep~\citep{pan2025introducingswegrepandswegrepmini} train specialized retrieval models with basic weighted $F_1$ rewards to issue fixed parallel calls, achieving speedups but requiring expensive inference infrastructure (Cerebras at 2800+ tokens/s). In web search, recent work trains models to distinguish parallelizable from sequential queries: HybridDeepSearcher~\citep{ko2025hybriddeepsearcherintegrating} fine-tunes models on synthetic hybrid-hop QA data to reduce search turns, while ParallelSearch~\citep{zhao2025parallelsearchtrainllmsdecompose} and RAG-R1~\citep{tan2025ragr1incentivizingsearchreasoning} apply RL with query decomposition for parallel execution.

However, existing work optimizes primarily for accuracy, leading to wasteful tool invocations that contribute nothing to search progress, or incomplete utilization of discovered information. While some approaches penalize trajectory length~\citep{zelikman2024quietstarlanguagemodelsteach}, this does not directly measure tool usage quality or distinguish between effective and redundant exploration.

We address this gap by introducing tool efficiency—the ratio of tools that discover new code entities to total tools invoked—as a metric that directly penalizes redundant exploration while encouraging focused information gathering. Our compact models achieve strong performance without specialized hardware by co-optimizing localization quality and search efficiency through efficiency-based RL rewards.

\section{Conclusion}

We introduced \textbf{FuseSearch}, a code localization agent that achieves superior accuracy-efficiency trade-offs through learned adaptive parallel execution. By optimizing \textbf{tool efficiency}—rewarding information novelty while penalizing redundancy—via SFT and RL, FuseSearch-4B achieves 84.7\% file-level $F_1$ while completing searches 93.6\% faster with 67.7\% fewer turns. As preprocessing for downstream repair tasks, it reduces interaction turns by 23.1\% and completion time by 28.5\%. Our results demonstrate that efficiency-aware training enables accurate yet computationally practical agents—a crucial step toward production-grade automated software development.

\section{Limitations}

Our work is subject to limitations in current evaluation frameworks. First, ground truth derived from golden patches represents only one valid solution, potentially missing alternative correct localizations. Second, available benchmarks predominantly cover Python repositories (SWE-bench Verified); while our toolset is language-agnostic, assessing effectiveness on statically-typed languages like Java or C++ requires more diverse benchmarks and training data. Third, existing benchmarks focus exclusively on issue-driven localization, whereas code search is fundamental to broader scenarios such as repository question answering, code comprehension, and documentation generation—contexts our approach has not been evaluated on. These constraints highlight the need for comprehensive localization benchmarks spanning diverse tasks and languages.

\bibliography{custom}

@misc{chen2025locagent,
  title={LocAgent: Graph-Guided LLM Agents for Code Localization}, 
  author={Zhaoling Chen and Xiangru Tang and Gangda Deng and Fang Wu and Jialong Wu and Zhiwei Jiang and Viktor Prasanna and Arman Cohan and Xingyao Wang},
  year={2025},
  eprint={2503.09089},
  archivePrefix={arXiv},
  primaryClass={cs.SE},
  url={https://arxiv.org/abs/2503.09089}
}

@misc{xia2024agentlessdemystifyingllmbasedsoftware,
      title={Agentless: Demystifying LLM-based Software Engineering Agents}, 
      author={Chunqiu Steven Xia and Yinlin Deng and Soren Dunn and Lingming Zhang},
      year={2024},
      eprint={2407.01489},
      archivePrefix={arXiv},
      primaryClass={cs.SE},
      url={https://arxiv.org/abs/2407.01489}, 
}

@misc{jiang2025issuelocalizationllmdriveniterative,
      title={Issue Localization via LLM-Driven Iterative Code Graph Searching}, 
      author={Zhonghao Jiang and Xiaoxue Ren and Meng Yan and Wei Jiang and Yong Li and Zhongxin Liu},
      year={2025},
      eprint={2503.22424},
      archivePrefix={arXiv},
      primaryClass={cs.SE},
      url={https://arxiv.org/abs/2503.22424}, 
}

@misc{ma2025toolintegratedreinforcementlearningrepo,
      title={Tool-integrated Reinforcement Learning for Repo Deep Search}, 
      author={Zexiong Ma and Chao Peng and Qunhong Zeng and Pengfei Gao and Yanzhen Zou and Bing Xie},
      year={2025},
      eprint={2508.03012},
      archivePrefix={arXiv},
      primaryClass={cs.SE},
      url={https://arxiv.org/abs/2508.03012}, 
}

@misc{qwen3technicalreport,
      title={Qwen3 Technical Report}, 
      author={Qwen Team},
      year={2025},
      eprint={2505.09388},
      archivePrefix={arXiv},
      primaryClass={cs.CL},
      url={https://arxiv.org/abs/2505.09388}, 
}

@misc{suresh2025cornstackhighqualitycontrastivedata,
      title={CoRNStack: High-Quality Contrastive Data for Better Code Retrieval and Reranking}, 
      author={Tarun Suresh and Revanth Gangi Reddy and Yifei Xu and Zach Nussbaum and Andriy Mulyar and Brandon Duderstadt and Heng Ji},
      year={2025},
      eprint={2412.01007},
      archivePrefix={arXiv},
      primaryClass={cs.CL},
      url={https://arxiv.org/abs/2412.01007}, 
}

@inproceedings{
    jimenez2024swebench,
    title={{SWE}-bench: Can Language Models Resolve Real-world Github Issues?},
    author={Carlos E Jimenez and John Yang and Alexander Wettig and Shunyu Yao and Kexin Pei and Ofir Press and Karthik R Narasimhan},
    booktitle={The Twelfth International Conference on Learning Representations},
    year={2024},
    url={https://openreview.net/forum?id=VTF8yNQM66}
}

@misc{wang2025openhandsopenplatformai,
      title={OpenHands: An Open Platform for AI Software Developers as Generalist Agents}, 
      author={Xingyao Wang and Boxuan Li and Yufan Song and Frank F. Xu and Xiangru Tang and Mingchen Zhuge and Jiayi Pan and Yueqi Song and Bowen Li and Jaskirat Singh and Hoang H. Tran and Fuqiang Li and Ren Ma and Mingzhang Zheng and Bill Qian and Yanjun Shao and Niklas Muennighoff and Yizhe Zhang and Binyuan Hui and Junyang Lin and Robert Brennan and Hao Peng and Heng Ji and Graham Neubig},
      year={2025},
      eprint={2407.16741},
      archivePrefix={arXiv},
      primaryClass={cs.SE},
      url={https://arxiv.org/abs/2407.16741}, 
}

@misc{yang2024sweagentagentcomputerinterfacesenable,
      title={SWE-agent: Agent-Computer Interfaces Enable Automated Software Engineering}, 
      author={John Yang and Carlos E. Jimenez and Alexander Wettig and Kilian Lieret and Shunyu Yao and Karthik Narasimhan and Ofir Press},
      year={2024},
      eprint={2405.15793},
      archivePrefix={arXiv},
      primaryClass={cs.SE},
      url={https://arxiv.org/abs/2405.15793}, 
}

@misc{zhao2025parallelsearchtrainllmsdecompose,
      title={ParallelSearch: Train your LLMs to Decompose Query and Search Sub-queries in Parallel with Reinforcement Learning}, 
      author={Shu Zhao and Tan Yu and Anbang Xu and Japinder Singh and Aaditya Shukla and Rama Akkiraju},
      year={2025},
      eprint={2508.09303},
      archivePrefix={arXiv},
      primaryClass={cs.CL},
      url={https://arxiv.org/abs/2508.09303}, 
}

@misc{ko2025hybriddeepsearcherintegrating,
      title={Hybrid Deep Searcher: Integrating Parallel and Sequential Search Reasoning}, 
      author={Dayoon Ko and Jihyuk Kim and Haeju Park and Sohyeon Kim and Dahyun Lee and Yongrae Jo and Gunhee Kim and Moontae Lee and Kyungjae Lee},
      year={2025},
      eprint={2508.19113},
      archivePrefix={arXiv},
      primaryClass={cs.AI},
      url={https://arxiv.org/abs/2508.19113}, 
}

@misc{tan2025ragr1incentivizingsearchreasoning,
      title={RAG-R1: Incentivizing the Search and Reasoning Capabilities of LLMs through Multi-query Parallelism}, 
      author={Zhiwen Tan and Jiaming Huang and Qintong Wu and Hongxuan Zhang and Chenyi Zhuang and Jinjie Gu},
      year={2025},
      eprint={2507.02962},
      archivePrefix={arXiv},
      primaryClass={cs.CL},
      url={https://arxiv.org/abs/2507.02962}, 
}

@misc{pan2025introducingswegrepandswegrepmini,
  author = {Ben Pan and Carlo Baronio and Albert Tam and Pietro Marsella and Mokshit Jain and Daniel Chiu and Swyx and Silas Alberti},
  year = {2025},
  url = {https://cognition.ai/blog/swe-grep},
  urldate = {October 16, 2025},
  title = {Introducing SWE-grep and SWE-grep-mini: RL for Multi-Turn, Fast Context Retrieval}
}

@misc{zelikman2024quietstarlanguagemodelsteach,
      title={Quiet-STaR: Language Models Can Teach Themselves to Think Before Speaking}, 
      author={Eric Zelikman and Georges Harik and Yijia Shao and Varuna Jayasiri and Nick Haber and Noah D. Goodman},
      year={2024},
      eprint={2403.09629},
      archivePrefix={arXiv},
      primaryClass={cs.CL},
      url={https://arxiv.org/abs/2403.09629}, 
}

@misc{shao2024deepseekmathpushinglimitsmathematical,
      title={DeepSeekMath: Pushing the Limits of Mathematical Reasoning in Open Language Models}, 
      author={Zhihong Shao and Peiyi Wang and Qihao Zhu and Runxin Xu and Junxiao Song and Xiao Bi and Haowei Zhang and Mingchuan Zhang and Y. K. Li and Y. Wu and Daya Guo},
      year={2024},
      eprint={2402.03300},
      archivePrefix={arXiv},
      primaryClass={cs.CL},
      url={https://arxiv.org/abs/2402.03300}, 
}

@inproceedings{kwon2023efficient,
  title={Efficient Memory Management for Large Language Model Serving with PagedAttention},
  author={Woosuk Kwon and Zhuohan Li and Siyuan Zhuang and Ying Sheng and Lianmin Zheng and Cody Hao Yu and Joseph E. Gonzalez and Hao Zhang and Ion Stoica},
  booktitle={Proceedings of the ACM SIGOPS 29th Symposium on Operating Systems Principles},
  year={2023}
}

@misc{li2025rllmrelationaltablelearning,
      title={rLLM: Relational Table Learning with LLMs}, 
      author={Weichen Li and Xiaotong Huang and Jianwu Zheng and Zheng Wang and Chaokun Wang and Li Pan and Jianhua Li},
      year={2025},
      eprint={2407.20157},
      archivePrefix={arXiv},
      primaryClass={cs.AI},
      url={https://arxiv.org/abs/2407.20157}, 
}

@article{gao2025more,
  title={More with Less: An Empirical Study of Turn-Control Strategies for Efficient Coding Agents},
  author={Gao, Pengfei and Peng, Chao},
  journal={arXiv preprint arXiv:2510.16786},
  year={2025}
}

@article{sheng2024hybridflow,
  title   = {HybridFlow: A Flexible and Efficient RLHF Framework},
  author  = {Guangming Sheng and Chi Zhang and Zilingfeng Ye and Xibin Wu and Wang Zhang and Ru Zhang and Yanghua Peng and Haibin Lin and Chuan Wu},
  year    = {2024},
  journal = {arXiv preprint arXiv: 2409.19256}
}

\appendix

\section{Parallel Execution Framework}
\label{sec:parallel_implementation}

\subsection{Inference Algorithm}

Algorithm~\ref{alg:rollout} describes the inference process of FuseSearch. At each turn, the model generates an action $a$ that may contain multiple tool calls formatted as JSON objects. The Parse function extracts these tool calls $\{c_1, \ldots, c_n\}$, which are then executed concurrently by the environment. The aggregated observations $\{o_1, \ldots, o_n\}$ are appended to the trajectory before the next model invocation. This cycle continues until the model produces a final answer without tool calls.

\begin{algorithm}[H]
\caption{Inference Process of FuseSearch}
\label{alg:rollout}
\begin{algorithmic}[1]
\Require Query $q$, model $\mathcal{M}$, environment $\mathcal{E}$
\Ensure Localized files $\mathcal{F}$
\State Initialize trajectory $\tau \leftarrow \emptyset$
\While{True}
    \State $a \leftarrow \mathcal{M}(q, \tau)$ // Generate action
    \State $\tau \leftarrow \tau \oplus a$ // Append action to trajectory
    \If{no \texttt{<tool\_call>} in $a$}
        \State \Return $answer \leftarrow$ Extract($a$)
    \EndIf
    \State $\{c_1, c_2, \ldots, c_n\} \leftarrow$ Parse($a$)
    \State $\{o_1, o_2, \ldots, o_n\} \leftarrow \mathcal{E}.\text{step}(\{c_1, \ldots, c_n\})$ 
    \State $\tau \leftarrow \tau \oplus \{o_1, o_2, \ldots, o_n\}$ 
\EndWhile
\end{algorithmic}
\end{algorithm}

\subsection{Tool Specifications}

Table~\ref{tab:tool_params} summarizes the core parameters of our three tools. Each tool is implemented as a function with a JSON-formatted parameter schema, enabling models to invoke them through structured tool calls.

\begin{table}[h]
  \centering
  \small
  \setlength{\tabcolsep}{4pt}
  \renewcommand{\arraystretch}{1.3}
  \begin{tabular}{lll}
    \hline
    \textbf{Tool} & \textbf{Required} & \textbf{Optional} \\
    \hline
    \texttt{read\_file} & \texttt{path} & \texttt{start\_line}, \texttt{end\_line} \\
    \texttt{grep} & \texttt{pattern} & \texttt{path}, \texttt{glob}, \texttt{output\_mode} \\
    \texttt{glob} & \texttt{pattern} & \texttt{path} \\
    \hline
  \end{tabular}
  \caption{Core parameters for the three minimalist tools. All paths are absolute.}
  \label{tab:tool_params}
\end{table}

\paragraph{read\_file} Reads file contents with optional line range specification. The \texttt{path} parameter specifies the absolute file path. When \texttt{start\_line} and \texttt{end\_line} are provided, only the specified range is returned; otherwise, the entire file is read (up to a default limit of 1000 lines).

\paragraph{grep} Performs regex-based content search built on ripgrep. The \texttt{pattern} parameter accepts full regex syntax. The optional \texttt{output\_mode} controls result format: \texttt{files\_with\_matches} (default) returns only file paths, \texttt{content} returns matching lines with context, and \texttt{count} returns match counts per file. The \texttt{glob} parameter filters files by pattern (e.g., \texttt{*.py}), while \texttt{path} restricts search to a specific directory.

\paragraph{glob} Matches files by name patterns. The \texttt{pattern} parameter accepts standard glob syntax (e.g., \texttt{**/*.js} for recursive search, \texttt{test\_*.py} for prefix matching). Results are limited to 100 file paths to prevent overwhelming context windows.

\section{Training Configuration}
\label{sec:training_config}

\subsection{Data Split}

From the collected repository-level localization dataset of approximately 21K issue-patch pairs, we allocate 6K samples for SFT and the remaining 15K samples for RL. The SFT subset undergoes trajectory synthesis and dual-metric filtering as described in Section~\ref{sec:training}, yielding approximately 6K high-quality demonstration trajectories.

\subsection{SFT}

We fine-tune Qwen3-4B-Instruct and Qwen3-30B-A3B-Instruct on the filtered trajectories for 1 epoch with a batch size of 32. Both models are trained on 8$\times$NVIDIA H20 GPUs(96GB). We use AdamW optimizer with a learning rate of 2e-5 and linear warmup for the first 10\% of training steps. The maximum sequence length is set to 32768 tokens to accommodate long repository contexts and multi-tool trajectories.

\subsection{RL}

\paragraph{Infrastructure} We employ vLLM~\citep{kwon2023efficient} as the inference engine to accelerate trajectory sampling during policy rollouts. The training framework is built on RLLM~\citep{li2025rllmrelationaltablelearning}, which leverages veRL~\citep{sheng2024hybridflow} for distributed RL.

\paragraph{Sampling Configuration} To ensure diversity in trajectory exploration, we set the sampling temperature to 0.7 during rollouts. For each training instance, we sample 8 trajectories (rollout=8) to compute advantage estimates for GRPO updates.

\paragraph{Training Hyperparameters} Table~\ref{tab:rl_hyperparams} summarizes the key hyperparameters for GRPO training. We use a per-GPU batch size of 32, yielding a global batch size of 256 across 32 NVIDIA H20 GPUs(96GB) with rollout factor 8. The prompt length is capped at 49152 tokens to accommodate extensive repository context, while response length is limited to 32768 tokens for tool call sequences and reasoning. We train for 1 epoch over the 15K RL training samples.

\begin{table}[h]
  \centering
  \small
  \setlength{\tabcolsep}{6pt}
  \renewcommand{\arraystretch}{1.3}
  \begin{tabular}{ll}
    \hline
    \textbf{Hyperparameter} & \textbf{Value} \\
    \hline
    Training batch size & 32 \\
    Rollout per instance & 8 \\
    Global batch size & 256 \\
    Sampling temperature & 0.7 \\
    Max prompt length & 49152 \\
    Max response length & 32768 \\
    Training epochs & 1 \\
    Training samples & 15K \\
    Learning rate & 1e-6 \\
    KL coefficient ($\beta$) & 0.01 \\
    \hline
  \end{tabular}
  \caption{Hyperparameters for GRPO-based RL.}
  \label{tab:rl_hyperparams}
\end{table}

\paragraph{Reward Coefficients} For the joint reward function $R(\tau) = \alpha \cdot F_1(\tau) + \gamma \cdot (F_1(\tau) \cdot e(\tau))$, we set $\alpha = 0.8$ and $\gamma = 0.2$. The $F_1$ score is computed as $F_1 = 0.7 \cdot F_1^{\text{file}} + 0.3 \cdot F_1^{\text{func}}$, placing higher weight on file-level accuracy while still incentivizing function-level precision.

\section{Baseline Implementation Details}
\label{sec:baseline_implementation}

We compare FuseSearch against four representative code localization frameworks, each employing distinct strategies and infrastructure requirements.

\paragraph{Agentless}~\citep{xia2024agentlessdemystifyingllmbasedsoftware} adopts a hierarchical pipeline approach without maintaining agent state across turns. It performs localization in three sequential stages: (1) file-level filtering using keyword matching and LLM ranking, (2) class/function identification within selected files, and (3) fine-grained line-level localization. This workflow-based design eliminates the need for complex reasoning chains but requires careful tuning of each pipeline stage.

\paragraph{CoSIL}~\citep{jiang2025issuelocalizationllmdriveniterative} implements an iterative agent that dynamically constructs module call graphs during exploration. Starting from entry points identified by keyword search, it progressively expands the graph by analyzing function invocations and dependencies. The agent employs context pruning to manage token limits, discarding less relevant code paths based on semantic similarity to the issue description. This dynamic graph construction enables adaptive exploration without requiring pre-built static analysis infrastructure.

\paragraph{LocAgent}~\citep{chen2025locagent} pre-processes repositories into directed heterogeneous graphs encoding file, class, and function nodes along with their structural relationships (imports, inheritance, invocations). The agent navigates this graph through node-visiting actions, leveraging graph connectivity to perform multi-hop reasoning. This approach requires upfront graph construction but enables efficient traversal of complex dependency chains.

\paragraph{RepoSearcher}~\citep{ma2025toolintegratedreinforcementlearningrepo} provides a lightweight tool suite (GetRepoStructure, SearchClass, SearchFunction, SearchClassMethod) for direct code retrieval without graph preprocessing. The agent iteratively invokes tools to gather relevant context, terminating via an explicit Exit action. This minimalist design reduces infrastructure overhead while maintaining competitive localization performance.

\subsection{Model Deployment}

For proprietary models (Claude-3.5-Sonnet, Kimi-K2-Instruct), we invoke their official APIs with temperature set to 0 for deterministic outputs. For open-source models (Qwen3-4B-Instruct, Qwen3-30B-A3B-Instruct), we deploy local inference servers using vLLM~\citep{kwon2023efficient} on 8$\times$NVIDIA H20 GPUs. We set tensor parallelism to 8 for distributed inference and enable continuous batching to maximize throughput. All models use their default system prompts and tool-calling formats as specified in their official documentation.

\section{Additional Experiments}
\label{sec:additional_experiments}

\subsection{Evaluation on LocBench}

To further validate the generalization of FuseSearch across different benchmarks, we conduct additional evaluation on LocBench~\citep{chen2025locagent}, a localization-focused benchmark designed to assess code localization capabilities. Following the same filtering criteria as SWE-bench Verified (excluding patches that introduce entirely new files or functions), we retain 456 out of 560 examples for evaluation.

Table~\ref{tab:locbench_results} presents the performance comparison between FuseSearch-4B (trained model) and Kimi-K2-Instruct on LocBench. Both models use the FuseSearch framework with parallel tool execution. The results demonstrate that our trained 4B model achieves superior localization quality while being significantly more efficient: it improves file-level F$_1$ by 5.4 points and function-level F$_1$ by 5.8 points compared to Kimi-K2, while reducing search time by 77\% (6.24s vs 27.8s) and token consumption by 35\% (37.5k vs 57.9k). The efficiency metric $e$ is also higher (74.06 vs 68.97), indicating that the trained model generates fewer redundant tool calls. These results confirm that our efficiency-aware training approach generalizes well beyond SWE-bench Verified, consistently producing models that achieve better quality-efficiency trade-offs across diverse localization tasks.

\begin{table}[h]
  \centering
  \small
  \setlength{\tabcolsep}{1.5pt}
  \renewcommand{\arraystretch}{1.3}
  \begin{tabular}{lccccccc}
    \hline
    \textbf{Model} & \textbf{File F$_1$} & \textbf{Func F$_1$} & \textbf{e} & \textbf{\#Turn} & \textbf{T(s)} & \textbf{Tok.(k)} \\
    \hline
    Kimi-K2 & 70.33 & 45.65 & 68.97 & 6.60 & 27.8 & 57.9 \\
    FuseSearch-4B & \textbf{75.74} & \textbf{51.47} & \textbf{74.06} & \textbf{4.61} & \textbf{6.24} & \textbf{37.5} \\
    \hline
  \end{tabular}
  \caption{Performance comparison on LocBench (456 examples). Both models use the FuseSearch framework with parallel execution.}
  \label{tab:locbench_results}
\end{table}

\section{Prompt Design}
\label{sec:prompt_design}

\subsection{System Prompt and Output Format}

\begin{figure*}[t]
  \centering
  \includegraphics[width=\textwidth]{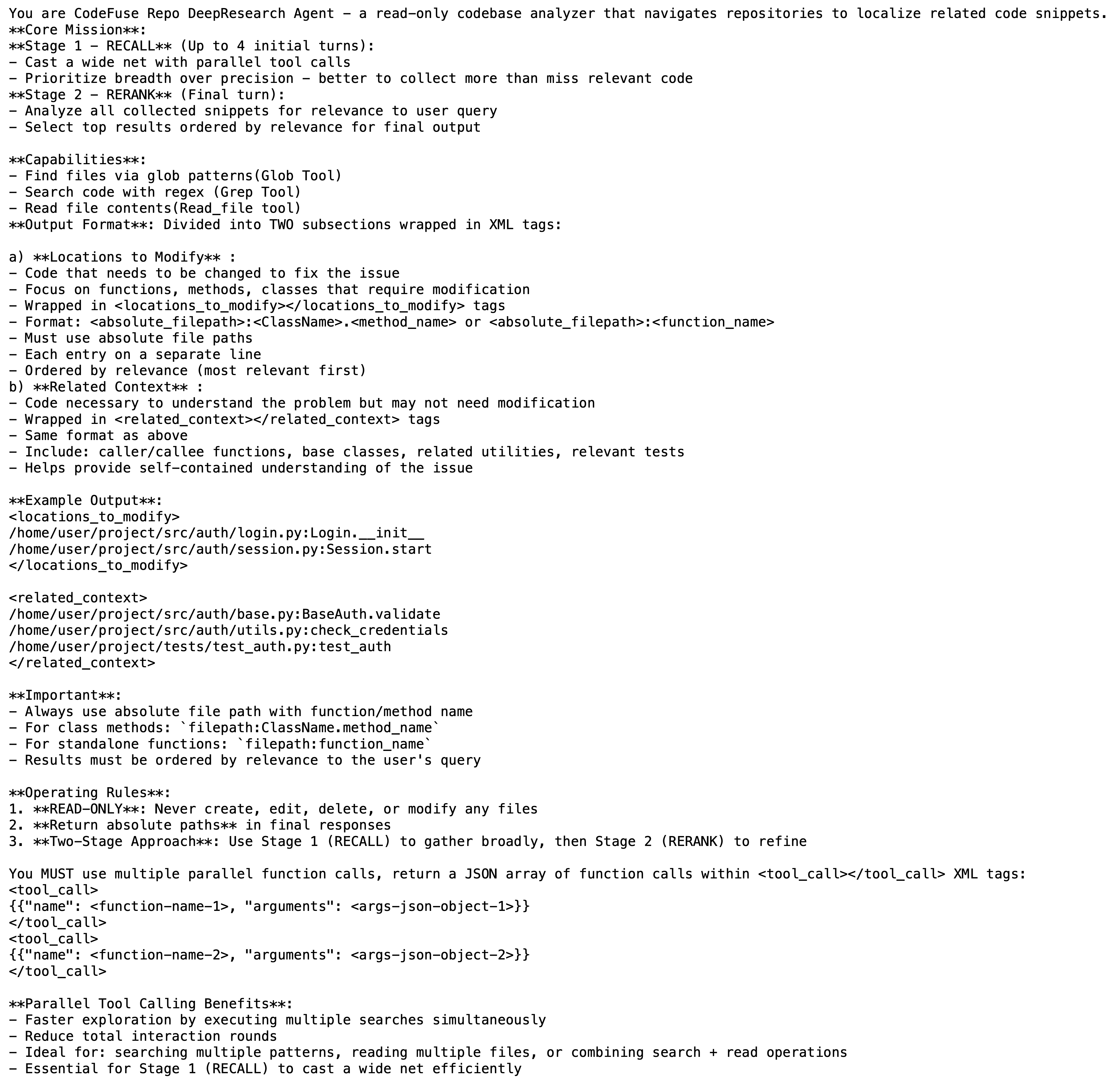}
  \caption{System prompt for FuseSearch. The prompt instructs the model to output localization results in two sections: Locations to Modify (required) and Related Context (optional).}
  \label{fig:system_prompt}
\end{figure*}

FuseSearch employs a structured prompt design that guides the model to produce localization results in two distinct sections: Locations to Modify and Related Context. Figure~\ref{fig:system_prompt} illustrates the complete system prompt used during inference.

\texttt{Locations to Modify} contains the core localization results—the specific files and functions that require modification to resolve the issue. The model outputs a ranked list of code entities, where higher-ranked items are deemed more likely to be the root cause. All precision, recall, and F$_1$ scores reported in our experiments are computed based on this section, treating it as the model's primary prediction.

\texttt{Related Context} allows the model to include additional code entities that are semantically related to the issue but do not necessarily require direct modification. For example, when localizing a bug in a data validation function, the model might include related utility functions or constants in Related Context even though the fix only requires modifying the validation logic itself. While not used for localization metric calculation, this supplementary information proves valuable for downstream issue resolution tasks: by providing repair agents with a broader context of relevant code, it enables faster and more accurate patch generation without requiring additional exploration turns.

This two-part output design reflects a key insight: effective localization should distinguish between "must-fix" locations (high precision for metrics) and "helpful context" (high utility for downstream tasks). By separating these concerns, FuseSearch simultaneously optimizes for localization accuracy and downstream task efficiency.

\end{document}